\pdfoutput=1

\documentclass[11pt]{article}

\usepackage[hyperref]{emnlp2021}
\usepackage{kotex} 
\usepackage{multirow}
\usepackage{algorithm}
\usepackage{algorithmic}

\usepackage{times}
\usepackage{latexsym}
\usepackage{graphicx}
\usepackage{amsmath}
\usepackage{amssymb}
\usepackage{booktabs}
\usepackage{adjustbox}
\usepackage{geometry}

\usepackage[T1]{fontenc}

\usepackage[utf8]{inputenc}

\usepackage{microtype}
\usepackage[T1]{fontenc}  
\usepackage{hyperref}    
\usepackage{url}      
\usepackage{booktabs}    
\usepackage{amsfonts}    
\usepackage{nicefrac}    
\usepackage{microtype}   
\usepackage{xcolor}     

\usepackage[normalem]{ulem}

\title{AVocaDo: Strategy for Adapting Vocabulary to Downstream Domain}

\author{Jimin Hong\thanks{~ Equal Contribution} , Taehee Kim\footnotemark[1] , Hyesu Lim\footnotemark[1] \and Jaegul Choo \\
    Korea Advanced Institute of Science and Technology (KAIST)\\
    \texttt{\{jimmyh, taeheekim, hyesulim, jchoo\}@kaist.ac.kr} \\
    }

\begin{document}
\maketitle
\begin{abstract}

During the fine-tuning phase of transfer learning, the pretrained vocabulary remains unchanged, while model parameters are updated.
The vocabulary generated based on the pretrained data is suboptimal for downstream data when domain discrepancy exists.
We propose to consider the vocabulary as an optimizable parameter, allowing us to update the vocabulary by expanding it with domain-specific  vocabulary based on a tokenization statistic.
Furthermore, we preserve the embeddings of the added words from overfitting to downstream data by utilizing knowledge learned from a pretrained language model with a regularization term.
Our method achieved consistent performance improvements on diverse domains (i.e., biomedical, computer science, news, and reviews).

\end{abstract}

\section{Introduction}
\label{sec:Intro}

A language model (LM) is pretrained with a large corpus in a general domain and then is fine-tuned to perform various downstream tasks, such as text classification, named entity recognition, and question answering.
However, fine-tuning the LM is challenging when the downstream domain is significantly different from the pretrained domain, requiring domain adaptation to improve the downstream performance~\citep{gururangan2020don, lee2020biobert, beltagy2019scibert}. 

Prior approaches conducted additional training with a large domain-specific corpus in between pretraining and fine-tuning.
In these approaches, the pretrained vocabulary remains unchanged, although the model is being adapted to a downstream domain, such as biomedicine or politics.

We argue that the vocabulary should also be adapted during the fine-tuning process towards downstream data.
Recent studies (e.g., SciBERT~\citep{beltagy2019scibert}) showed that using an optimized vocabulary for a particular downstream domain is more effective than using the vocabulary generated in pretraining stage.
However, these approaches required a large domain-specific corpus additional to the downstream data in order to construct optimized vocabulary for the downstream domain.

\begin{figure}[t] 
\includegraphics[width=0.49\textwidth]{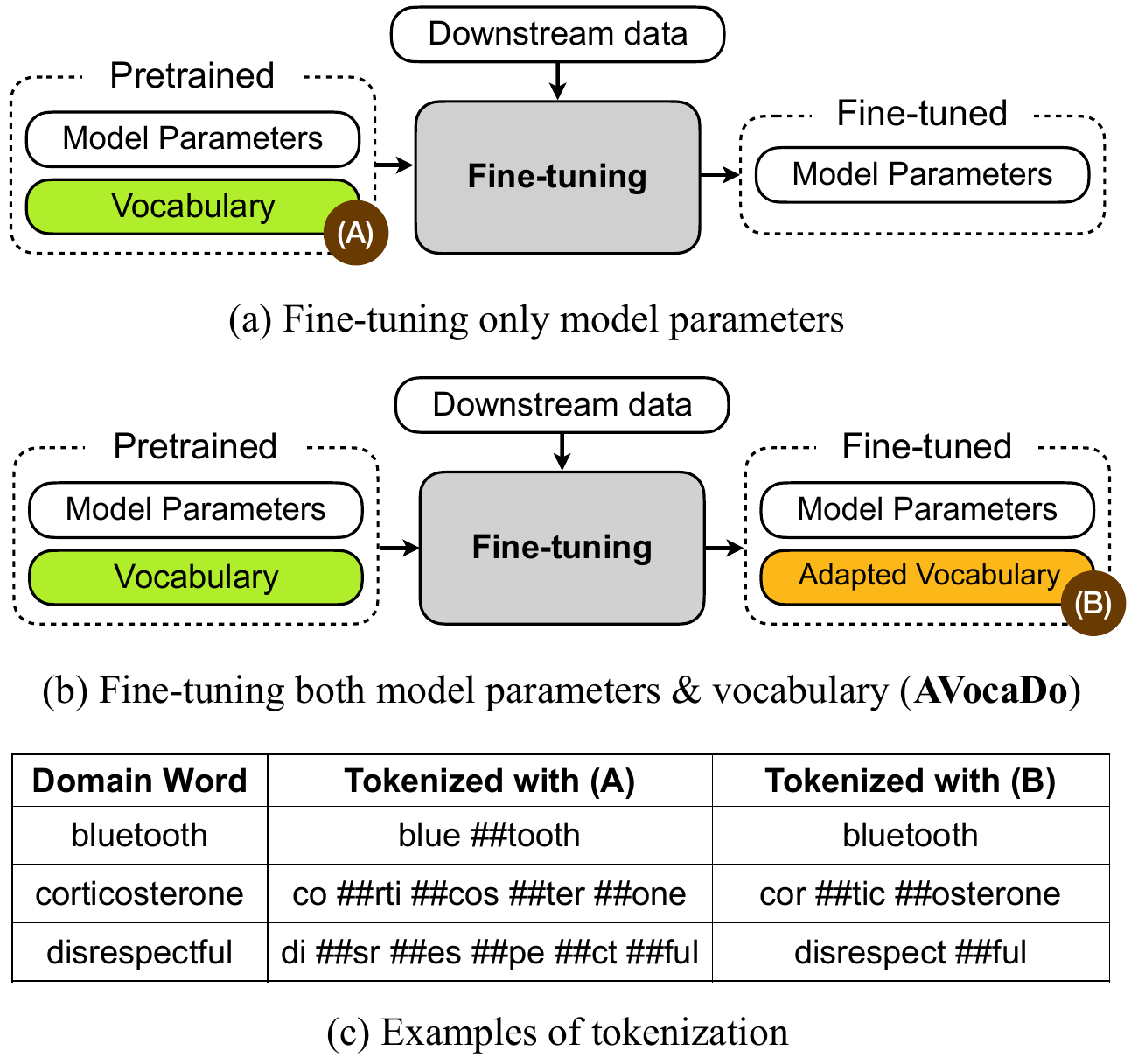}
\caption{\textbf{Overview of AVocaDo.} AVocaDo updates the vocabulary (b) not only fine-tuning the model parameters as done by previous approaches (a). 
fine-tuning the vocabulary has benefit on tokenizing domain-specific words (c).}
\label{fig:1_overview}
\end{figure}

We propose to \textbf{A}dapt the \textbf{Voca}bulary to downstream \textbf{Do}main (\textbf{AVocaDo}), which updates the pretrained vocabulary by expanding it with words from the downstream data without requiring additional domain-specific corpus. 
The relative importance of words is considered in determining the size of the added vocabulary.
As shown in Figure~\ref{fig:1_overview}-(c), domain-specific words are tokenized in unwilling manner in the corresponding domain. 
For example, in reviews domain, the "bluetooth" represents a short-range wireless technology standard, but when the word is tokenized into "blue" and "tooth", the combined meaning of each subword is totally different from the intended meaning of "bluetooth".
Furthermore, we propose a regularization term that prevents the embeddings of added words from overfitting to downstream data, since downstream data is relatively small compared to the pretraining data.

The experimental results show that our proposed method improves the overall performance in a wide variety of domains, including biomedicine, computer science, news, and reviews.
Moreover, the advantage of the domain adapted vocabulary over the original pretrained vocabulary is shown in qualitative results.

\section{Related Work}

As transfer learning has shown promising results in natural language processing (NLP), recent work leveraged the knowledge learned from the pretrained model, such as BERT~\citep{devlin2018bert} in various domains. 

SciBERT~\citep{beltagy2019scibert} trains a language model with the large domain-specific corpus from scratch, showing that the vocabulary constructed from the domain-specific corpus contributes to improving performance. 
\citet{lee2020biobert} and \citet{gururangan2020don} conducted additional training on a pretrained LM with a large domain-specific corpus before fine-tuning. 
On the other hand, exBERT~\citep{tai2020exbert} extended the pretrained model with new vocabulary to adapt to biomedical domain. Similarly, \citet{poerner2020inexpensive} and \citet{sato2020vocabulary} proposed to expand vocabulary and leverage external domain-specific corpus to train new embedding layers.

On the contrary, AVocaDo requires only downstream dataset in domain adaptation.
Furthermore, our method selects a subset of domain-specific vocabulary considering the relative importance of words.

\section{Methods}

In \textbf{AVocaDo}, we generate domain-specific vocabulary based on the downstream corpus. 
The subset of the generated vocabulary is merged with the original pretrained vocabulary. The size of subset is controlled by the fragment score.
Afterwards, we apply a regularization term during fine-tuning to prevent the embeddings of added words from overfitting to the downstream data.

\begin{algorithm}[tb] 
 \caption{Adapting Vocab in AVocaDo}
 \label{alg:merge vocab}
\begin{algorithmic}[1]
  
  \STATE \textbf{Input}: pretrained vocab $V_\mathcal P$; corpus $\bold C$;
              \\~~~~~~~~~~~~domain-specific vocab $V_\mathcal D$.
  \STATE \textbf{Output}: adapted vocab $V_\mathcal A$.
  \STATE Initialize hyperparamenters $\alpha, \beta, \gamma$.
  \STATE $V_\mathcal A \leftarrow V_\mathcal P \cup \{V_{\mathcal D_i}\}_{i=0}^{\alpha}$ 
  \WHILE{$ f_{\bold C}(V_\mathcal A) > \gamma $}
    \STATE $V_\mathcal A \leftarrow V_\mathcal A \cup \{V_{\mathcal D_i}\}_{i=\alpha}^{\alpha+\beta}$
    \STATE $\alpha \leftarrow \alpha+\beta$
  \ENDWHILE
  \RETURN $V_\mathcal A$
\end{algorithmic}
\end{algorithm}

\begin{figure*}[t] 
\includegraphics[width=1\textwidth]{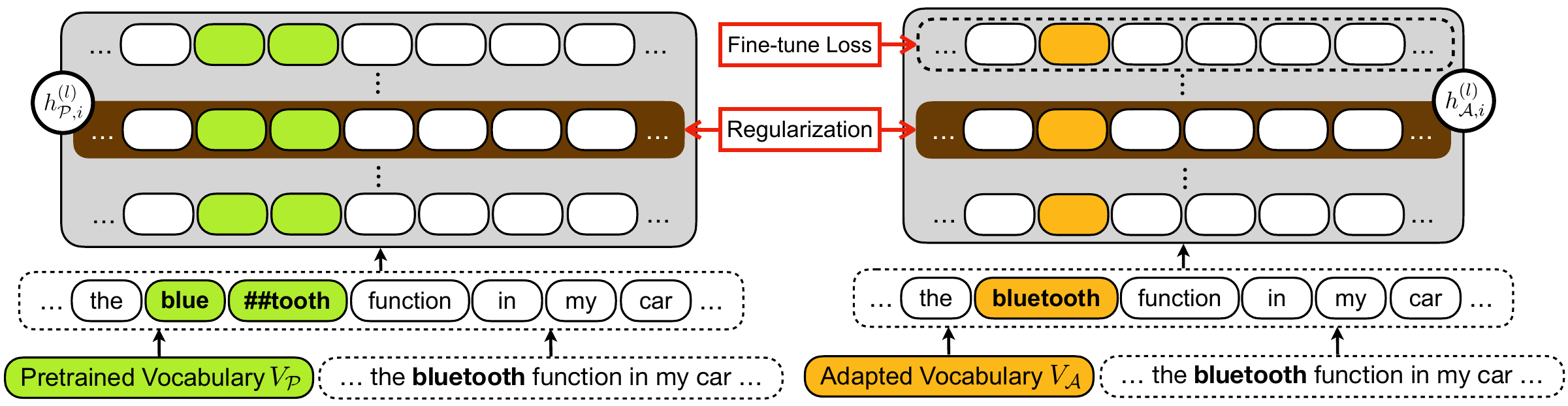}
\caption{
\textbf{Fine-tuning with regularization.} 
Identical sentence "... the bluetooth function in my car ...", sampled from $\textsc{Amazon}$, is tokenized with pretrained vocabulary (left) and with adapted vocabulary (right).
The domain-specific word "bluetooth" is tokenized in two ways, which are highlighted as green and yellow respectively.
The model is fine-tuned with regularization on $l$-th layer, highlighted as brown box, to preserve the embeddings of added words (e.g., bluetooth) from overfitting to downstream dataset.
}
\label{fig:2_align}
\vspace{-0.4cm}
\end{figure*}

\subsection{Adapting Vocabulary}
\label{method:autovocab} 

In this section, we describe the procedure of adapting the vocabulary to the downstream domain through Algorithm~\ref{alg:merge vocab}.
First, the domain-specific vocabulary set $V_{\mathcal D}$ is constructed from the downstream corpus $\bold C$ given a vocabulary size $N_{\mathcal D}$ and a tokenizing algorithm.
The adapted vocabulary set $V_{\mathcal A}$ is constructed by merging the subset of $V_{\mathcal D}$, size of $n_{\mathcal D}$, with the original pretrained vocabulary set $V_{\mathcal P}$, size of $N_{\mathcal P}$.
In other words, $N_{\mathcal A}$, the size of $V_{\mathcal A}$, is equal to the sum of the merged vocabulary sets, i.e, $N_{\mathcal A} = n_{\mathcal D}+N_{\mathcal P}$.
Note that $n_{\mathcal D} < N_{\mathcal D}$, because when too many words are added, the added infrequent subwords might cause the rare word problem~\citep{luong2014addressing, schick2020rare}.

The subset of $V_{\mathcal D}$, that is added to $V_{\mathcal P}$, is determined by the \textit{fragment score} $f_{\bold C}(V)\in \mathbb R$, which we introduce as a new metric that measures the relative number of subwords tokenized by a vocabulary $V$ from a single word in corpus $\bold C$, i.e., 
\begin{equation}
\begin{adjustbox}{max width=\textwidth}
  $f_{\bold C}(V) = \frac{\text{the number of subwords tokenized by }V}{\text{the number of words in }\bold C}.$
\end{adjustbox}
\end{equation}

Motivated by~\citet{goodtokenizer21}, we keep $f_{\bold C}(V_{\mathcal A})$ from exceeding a certain threshold $\gamma$.
$\gamma$ is a hyperparameter determining the lower bound of the $f_{\bold C}(V_{\mathcal A})$.
Decreasing the lower bound leads $V_{\mathcal A}$ to less finely tokenize $\bold C$. 
In contrast, increasing the lower bound leads $V_{\mathcal A}$ to finely tokenize $\bold C$.

We sought to consider the importance of subwords when adding $V_{\mathcal D}$. 
We simply selected a subset of $V_{\mathcal D}$ following the order of merging subwords used in byte pair encoding algorithm~\citep{sennrich2015neural}.
The number of added vocabulary in each iteration is indicated by the hyperparameters $\alpha$ and $\beta$.

In summary, as frequent subword pairs are added from $V_{\mathcal D}$ to $V_{\mathcal A}$ as subwords, the $f_{\bold C}(V_{\mathcal A})$ decreases.
The objective of adding $V_{\mathcal D}$ to $V_{\mathcal A}$ is to decrease the $f_{\bold C}(V_{\mathcal A})$, but we make sure that $f_{\bold C}(V_{\mathcal A})$ does not become too small, i.e., lower than the threshold $\gamma$.
Therefore, we continue to add $V_{\mathcal D}$ to $V_{\mathcal A}$ if $f_{\bold C}(V_{\mathcal A})$ is higher than $\gamma$, and terminate the merging step otherwise.

\subsection{Fine-tuning with Regularization}
\label{method:contrastive}

The embeddings of words in the subset of $V_{\mathcal D}$ which is merged with $V_{\mathcal P}$ to construct the adapted vocabulary $V_{\mathcal A}$ are trained only with downstream data during fine-tuning.
Since the size of downstream data is much smaller than that of the pretraining corpus, 
the embeddings trained only with the downstream data possibly suffer from overfitting.
To prevent the potential overfitting, we leverage the pretrained contextual representation learned from a large corpus.

In contrastive learning~\citep{simclr20}, a pair of instances is encouraged to learn representations in relation to the similarity of the instances. 
We apply this contrastive learning framework as a regularization in fine-tuning. 
As described in Figure~\ref{fig:2_align}, an identical sentence is tokenized in two ways: one with the pretrained vocabulary $V_{\mathcal P}$ and the other with the adapted vocabulary $V_{\mathcal A}$. 
A minibatch consists of $B$ input sentences $\bold x = \{x_1, \ldots x_{B}\}$.
Each input $x_i$ is tokenized with two types of vocabularies, and their $l$-th layer encoder outputs are denoted as $h_{\mathcal P, i}^{(l)}$ and $h_{\mathcal A, i}^{(l)}$.
Note that they are encoded with a single encoder given the identical input sentence, but with different tokenizations.
$h_{\mathcal P, i}^{(l)}$ and $h_{\mathcal A, j}^{(l)}$ are considered as a positive pair when $i=j$, and as a negative pair when $i\neq j$.
The positive pair $h_{\mathcal P, i}^{(l)}$ and $h_{\mathcal A, i}^{(l)}$ are trained to maximize the agreement by the regularization term $\mathcal L_{\text{reg}}$ i.e., 

\small
\begin{equation}
\label{eq:align_loss}
\begin{aligned}
\setlength\textwidth{7cm}
  &\mathcal{L}_\text{reg} (\bold h_{\mathcal A}^{(l)}, \bold h_{\mathcal P}^{(l)}) \\
  &= -\dfrac{1}{B}
   \log\sum_{i=1}^B \frac {e^{(\text{sim}(h_{\mathcal A, i}^{(l)}, h_{\mathcal P, i}^{(l)})/\tau)}}
        {\sum_{j=1}^B e^{(\text{sim}(h_{\mathcal A, i}^{(l)}, h_{\mathcal P, j}^{(l)})/\tau)}},
\end{aligned}
\end{equation}
\normalsize
where $\tau$ is a softmax temperature, $B$ is a batch size, $\bold h_{\mathcal A}^{(l)}=\{h_{\mathcal A, 1}^{(l)}, \ldots, h_{\mathcal A, B}^{(l)}\}$ and $\bold h_{\mathcal P}^{(l)} = \{h_{\mathcal P, 1}^{(l)},\ldots, h_{\mathcal P, B}^{(l)}\}$.
The cosine similarity function is used for $\text{sim}(\cdot)$.
$\bold h_{\mathcal A}^{(l)}$ is prevented from overfitting by making it closer to its positive sample.

The model is trained to perform the target task with the regularization term $\mathcal L_{reg}$.
The output of the encoder with $V_{\mathcal{A}}$ is supervised by the label of downstream data with cross entropy loss $\mathcal L_{CE}$.
The total loss $\mathcal L$ for domain adaptive fine-tuning is formalized as

\small
\begin{equation}
\begin{aligned}
  \mathcal L &= \mathcal L_{CE} + \lambda\mathcal L_{reg},\\
  \mathcal L_{CE} &= -\frac{1}{B}\sum^{B}\sum_{i=1}^{C} t_{i}\log(f(s_{i})),
\end{aligned}
\end{equation}
\normalsize
where $f$ is a softmax function, $C$ is the total number of classes, $s_{i}$ is the logit for $i$-th class, $B$ is the batch size, and $t_{i}$ is the target label. 
In our implementation, we set $\lambda$ as 1.0 for all experiments.

\begin{table*}[t]
\centering 
  \begin{adjustbox}{width=1\textwidth}
    \begin{tabular}{l| l | c c| cc|cc} 
    \toprule
    \textbf{Domain} &\textbf{Dataset} &\textbf{BERT}$_{\text{base}}$ &$\textbf{BERT}_{
    \text{AVocaDo}}$ &$\textbf{SciBERT}$ &$\textbf{SciBERT}_{
    \text{AVocaDo}}$ & \textbf{BioBERT} & $\textbf{BioBERT}_{
    \text{AVocaDo}}$
    \\ [0.1ex] 
    \midrule
    \textsc{BioMed} &\textsc{ChemProt}  &$79.38$ &$\textbf{81.07}(+1.69)$ &$82.16$ &$\textbf{82.71}(+0.55)$ &$83.58$ &$\textbf{84.42}(+0.84)$ 
    \\ [0.1ex]
    \hline  
    CS 
      &ACL-ARC &$56.82$ &$\textbf{67.28}(+10.46)$ &$66.89$ &$\textbf{75.02}(+8.13)$ &- 
    \\ [0.1ex] 
    \hline  
    \textsc{News} 
    &\textsc{HyperPartisan} &$84.51$ &$\textbf{89.31}(+4.80)$ &- &- &- &- 
    \\ [0.1ex] 
    \hline  
    \textsc{Reviews} &\textsc{Amazon} &$55.50$ &$\textbf{68.51}(+13.01)$ &- &- &- &- 
    \\ [0.1ex] 
    
    \bottomrule 
    \end{tabular}
  \end{adjustbox}
  \caption{\textbf{Comparisons with baselines in four different domains.} 
  Pretrained LMs (i.e., BERT$_{\text{base}}$, SciBERT, and BioBERT) are fine-tuned in two ways: one with pretrained vocabulary (represented without a subscription) and the other with adapted vocabulary (represented with subscription AVocaDo).
  The performance improvement is represented inside the parentheses with $+$. The reported value is averaged $F_{1}$ score (micro-$F_{1}$ for \textsc{ChemProt} and macro-$F_{1}$ for the others) over five random seeds. Invalid comparisons are represented as -.} 
  \label{tab:main_result}
\end{table*}

\section{Experimental Settings}
\noindent\textbf{Datasets} 
We conducted experiments on four domains that are significantly different from the pretraining domain; biomedical (\textsc{BioMed}) papers, computer science (\textsc{CS}) papers, \textsc{News}, and amazon reviews (\textsc{Reviews}). 
\textsc{ChemProt}~\citep{10.1093/database/bav123}, ACL-ARC~\citep{jurgens-etal-2018-measuring}, \textsc{HyperPartisan}~\citep{kiesel-etal-2019-semeval}, and \textsc{Amazon}~\citep{mcauley2015image} datasets are used in respective domains.
Target task for each dataset is text classification. Appendix~\ref{appendix:implementation details} describes more details.

\noindent\textbf{Evaluation Protocol  } 
We report the macro-$F_{1}$ score for ACL-ARC, \textsc{HyperPartisan}, and \textsc{Amazon} and micro-$F_{1}$ score for \textsc{ChemProt} as done by previous work~\citep{lee2020biobert,beltagy2019scibert}.
The score is averaged over five random seeds.


\section{Results}
\subsection{Quantitative Results}
BERT$_{\text{base}}$~\citep{devlin2018bert}, SciBERT~\citep{beltagy2019scibert}, and BioBERT~\citep{lee2020biobert} are chosen as the pretrained LMs for our experiments.
Each model is fine-tuned in two ways: one with pretrained vocabulary and the other with adapted vocabulary. 

SciBERT is pretrained with scientific corpus while BERT is pretrained with general domain corpus (e.g., Wikipedia), and thus SciBERT can be fine-tuned only with $\textsc{BioMed}$ and $\textsc{CS}$.
BioBERT conducted additional training with biomedical corpus, so that BioBERT can be fine-tuned only with $\textsc{BioMed}$.

As described in Table~\ref{tab:main_result}, fine-tuning with AVocaDo significantly improved the performance of the downstream task in all domains.
Note that the performance is improved despite the low-resource environment, where the size of dataset is smaller than 5,000 as described in Appendix~\ref{appendix:implementation details} (\textsc{ChemProt}, \textsc{ACL-ARC}, and \textsc{HyperPartisan}).
In \textsc{BioMed} domain, applying AvocaDo improved the overall performance in various pretrained language models. 
This improvement shows that utilizing the domain-specific vocabulary has additional benefits on the downstream domain. 
In \textsc{CS}, AVocaDo outperforms BERT$_{\text{base}}$ and SciBERT, showing the performance improvements of 10.46 in BERT$_{\text{base}}$ and 8.13 in SciBERT.
In \textsc{News} and \textsc{Reviews}, our strategy significantly improved the performance; 4.80 in \textsc{News} and 13.01 in \textsc{Reviews}.

\begin{table}[t]
\begin{adjustbox}{width=0.48\textwidth}
\begin{tabular}{c|lll}
\toprule
\textbf{Domain}                                              & \multicolumn{1}{c}{\textbf{Domain Word}} & \multicolumn{1}{c}{\textbf{Pretrained Vocab $V_\mathcal{P}$}} & \multicolumn{1}{c}{\textbf{Adapted Vocab $V_\mathcal{A}$}} \\ \midrule
\multicolumn{1}{c|}{\multirow{2}{*}{\textsc{BioMed}}}                        & glucuronidation     & g, lu, cu, ron, ida, tion & glucuron, ida, tion    \\
\multicolumn{1}{c|}{}                                      & sulfhydration      & sul, f, hy, dra, tion   & sulf, hydr, ation     \\ \hline
\multicolumn{1}{c|}{\multirow{2}{*}{\begin{tabular}[c]{@{}c@{}}\textsc{CS}\end{tabular}}} & nlp$^{*}$           & nl, p           & nlp            \\
\multicolumn{1}{c|}{}                                      & syntactic        & syn, ta, ctic       & syntactic         \\ \hline
\multicolumn{1}{c|}{\multirow{2}{*}{\textsc{News}}}                            & tweet          & t, wee, t         & tweet           \\
\multicolumn{1}{c|}{}                                      & disrespectful      & di, sr, es, pe, ct, ful  & disrespect, ful      \\ \hline
\multicolumn{1}{c|}{\multirow{2}{*}{\textsc{Reviews}}}                          & otterbox         & otter, box        & otterbox          \\
\multicolumn{1}{c|}{}                                      & thunderbolt       & thunder, bolt       & thunderbolt        \\ 
\bottomrule
\end{tabular}
\end{adjustbox}
\caption{\textbf{Qualitative results.} Carefully selected tokenization examples from $V_\mathcal{P}$ and $V_\mathcal{A}$. $*$ represents capitalized in the original sentence.}
\label{tab:qualitative}
\vspace{-0.4cm}
\end{table}

\subsection{Qualitative Results}
\label{sec:qualitative results}
To analyze the effectiveness of the adapted vocabulary $V_{\mathcal A}$, we show the sampled words from each domain that are tokenized with two types of vocabulary in Table~\ref{tab:qualitative}. 

The adapted vocabulary $V_{\mathcal A}$ tokenizes the domain-specific word into subwords that are informative in the target domain.
For example, in the case of "sulfhydration", the word is tokenized as "sul, f, hy, dra, tion" with $V_{\mathcal{P}}$ and "sulf, hydr, ation" with $V_{\mathcal{A}}$.
"sulf" and "hydr" imply "sulfur" and "water" respectively, which are frequently used in \textsc{BioMed} domain. 

Furthermore, $V_{\mathcal A}$ preserves the semantic of a domain-specific word by keeping it as a whole word, where the subwords tokenized with $V_{\mathcal P}$ have completely different semantics from its original meaning.
For instance, "otterbox" is an electronics accessory company in the \textsc{Reviews} domain.
However, with $V_{\mathcal P}$, it is split into "otter" and "box", where the "otter" is a carnivorous mammal and "box" is a type of container. 
Randomly sampled tokenization examples from $V_\mathcal{P}$ and $V_\mathcal{A}$ are presented in Appendix Table~\ref{tab:appendix:random_qualitative}.


\begin{table}[t]
  \begin{adjustbox}{width=0.49\textwidth}
    \begin{tabular}{l@{\hskip6pt}|@{\hskip6pt}c@{\hskip6pt}c@{\hskip6pt}c@{\hskip6pt}c}
      \toprule
      \textbf{Model} & \textsc{ChemProt} & \textsc{ACL-ARC} & \textsc{HyperPartisan} & \textsc{Amazon} \tabularnewline \midrule
      AVocaDo & 81.07 & 67.28 & 89.31 &68.51 \tabularnewline \hline
      w/o $\mathcal{L}_{reg}$ & 78.45(-2.62) & 64.00(-3.28) & 87.84(-1.47) & 61.23(-7.28) \tabularnewline \hline
      BERT$_{\text{base}}$ &79.35(-1.72) &56.82(-10.46) &84.51(-4.80) & 55.50(-13.01)   \tabularnewline \bottomrule

    \end{tabular}
\end{adjustbox}
\caption{\textbf{Ablation study.}
  w/o $\mathcal{L}_{reg}$ denotes that the model is fine-tuned with the adapted vocabulary but not applying regularization loss.
  BERT$_\text{base}$ denotes that the model is fine-tuned without applying AVocaDo. The performance difference is represented inside the parentheses.}
\label{tab:ablation}
\vspace{-0.4cm}
\end{table}

\subsection{Ablation Studies}
The effectiveness of each component in AVocaDo, i.e., vocabulary adaptation and contrastive regularization, is shown in this section.
As described in Table \ref{tab:ablation}, vocabulary adaptation improves the performance in three domains (i.e., \textsc{ACL-ARC}, \textsc{HyperPartisan}, and \textsc{Amazon}) even in the absence of the regularization term.

\subsection{Size of Added Vocabulary}

The size of the added vocabulary $n_{\mathcal D}$ is automatically determined by the fragment score of the adapted vocabulary $V_{\mathcal A}$, as described in Algorithm~\ref{alg:merge vocab}. 
In order to analyze how $n_{\mathcal D}$ affects the performance, we compare the performance of downstream tasks by manually setting the $n_{\mathcal D}$ as 500, 1000, 2000, and 3000 without using the fragment score, as shown in Table~\ref{tab:size of added vocab}.
Automatically determined $n_{\mathcal D}$ is 1600, 700, 2850 and 1300 for each dataset. 
Except for \textsc{Amazon} dataset, we demonstrate that determining $n_{\mathcal D}$ by the fragment score shows the optimal performance.

\section{Conclusion}
In this paper, we demonstrate that a pretrained vocabulary should be updated towards a downstream domain when fine-tuning. 
We propose a fine-tuning strategy called \textbf{AVocaDo} that adapts the vocabulary to the downstream domain by expanding the vocabulary based on a tokenization statistic, and by regularizing the newly added words. 
Our approach shows consistent performance improvements in diverse domains on various pretrained language models. 
AVocaDo is applicable to a wide range of NLP tasks in diverse domains without any restrictions, such as massive computing resources or a large domain-specific corpus.

\begin{table}[h]
  \begin{adjustbox}{width=0.49\textwidth}
    \begin{tabular}{l|ccccc}
      \toprule
      & \multicolumn{5}{c}{\textbf{Size of added vocabulary $n_{\mathcal D}$}} \tabularnewline
      \multicolumn{1}{c|}{\textbf{Dataset}} & \multicolumn{1}{c}{500}
      & \multicolumn{1}{c}{1000} & \multicolumn{1}{c}{2000}
      & \multicolumn{1}{c}{3000} & \multicolumn{1}{c}{AVocaDo}
      \\ \midrule
      \textsc{ChemProt} &
      80.36 &	80.43 & 80.24 &	79.89 &	\textbf{81.06} \\ \hline
      \textsc{ACL-ARC} &
      65.24 &	64.43 &	66.08 &	65.37 &	\textbf{67.28} \\ \hline
      \textsc{HyperPartisan} &
      84.70	&85.03	&80.49	&84.85 & \textbf{89.31}
      \\ \hline
      \textsc{Amazon} &
      68.57	&68.31 &\textbf{68.85}	&67.89 &68.51 
      \\ 
      \bottomrule
    \end{tabular}
\end{adjustbox}
\caption{\textbf{Analysis on the size of the added vocabulary.} $n_{\mathcal D}$ is manually set (500, 1000, 2000, and 3000) or automatically determined (AVocaDo).}
\label{tab:size of added vocab}
\vspace{-0.4cm}
\end{table}

\section*{Acknowledgment}
This work was supported by Institute of Information \& communications Technology Planning \& Evaluation (IITP) grant funded by the Korea government(MSIT) (No.2019-0-00075, Artificial Intelligence Graduate School Program(KAIST)), and Samsung Advanced Institute of Technology, Samsung Electronics Co., Ltd.
We thank the anonymous reviewers for their helpful feedback and discussions. We also thank Seong-Su Bae for his insight and helpful opinion.

\bibliography{references}
\bibliographystyle{plainnat}

\clearpage
\newpage

\appendix
\section*{Appendix}
\label{sec:appendix}

\begin{figure*}[h!] 
\centering
\includegraphics[width=1\textwidth]{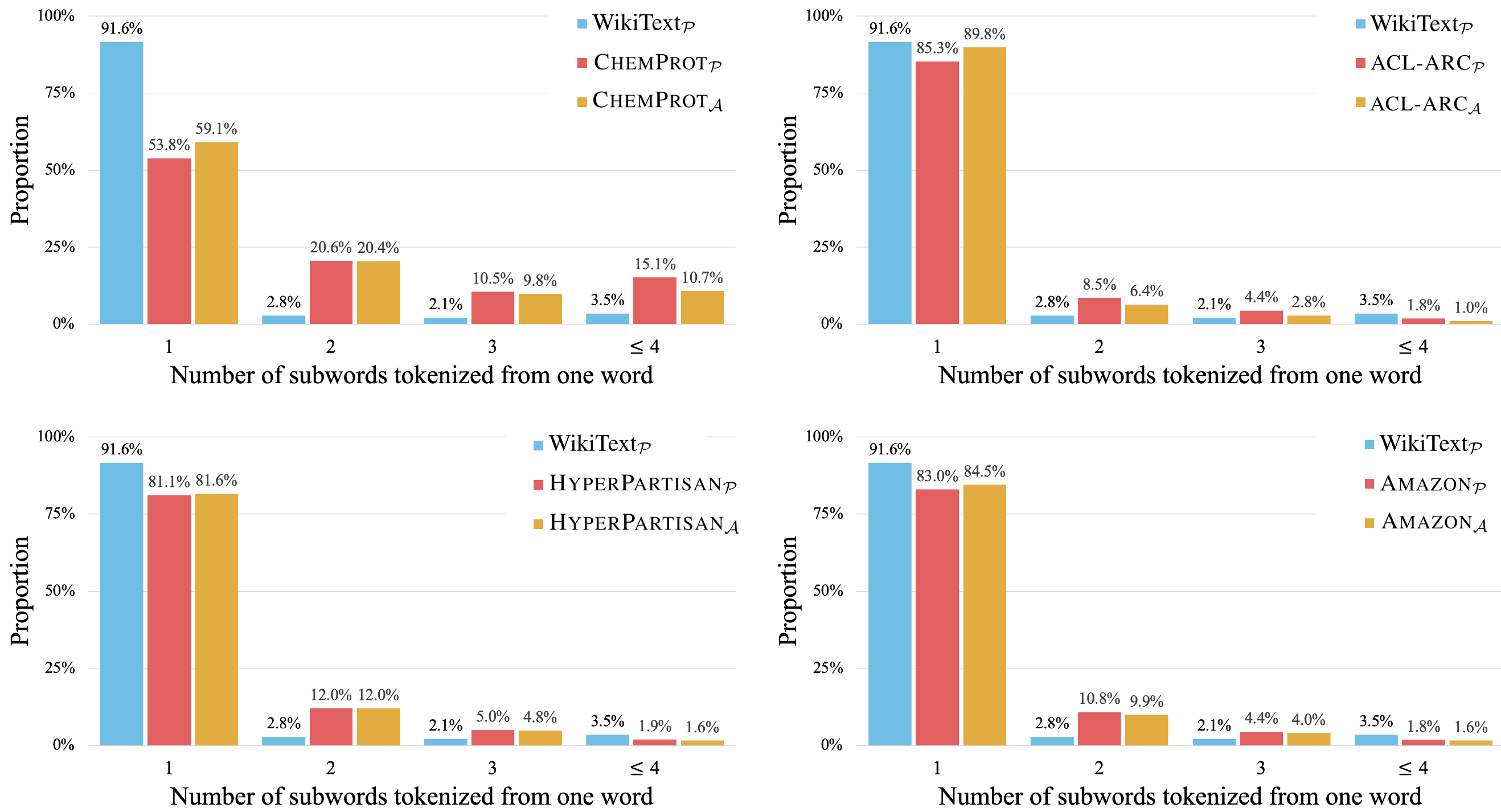}
\caption{The analysis of the pretrained and adapted vocabularies on WikiText and downstream domains. $\mathcal {P}$ and $\mathcal {A}$ denote the pretrained vocabulary and the adapted vocabulary respectively. \textbf{AVocaDo} mitigates the domain gap in terms of the average number of subwords tokenized from a single word.}
\label{fig:appendix:behavior of tokenizer}
\end{figure*}

\section{Details on Fragment Score}
\label{appendix:fragment_score}
Fragment score is a measure of the \textit{fineness} of tokenization.
We observed that the pretrained vocabulary set $V_{\mathcal P}$ tokenizes domain-specific words (i.e., words that are frequently appeared in a downstream corpus but not in a pretrained corpus) into larger number of subwords than the number of subwords that non-domain-specific words are tokenized into (Figure~\ref{fig:appendix:behavior of tokenizer}).
These finely tokenized subwords are not semantically informative enough.

Inspired by the observations, we construct a new vocabulary $V_{\mathcal A}$ that \textit{less finely} tokenizes the domain-specific words than $V_{\mathcal P}$, i.e., $V_{\mathcal A}$ such that $f_{\bold C}(V_{\mathcal A}) < f_{\bold C}(V_{\mathcal P})$.
This is why we chose the fragment score of the newly constructed vocabulary set $V_{\mathcal A}$ as a metric for selecting a subset of domain-specific vocabulary $V_{\mathcal D}$.

\section{Different Aspects of the Vocabularies}
\label{sec:appendix:vocab behabior}

Figure~\ref{fig:appendix:behavior of tokenizer} shows the relative number of tokenized subwords from a single word in four domains where the publicly available vocabulary in BERT~\citep{devlin2018bert} is denoted as $V_{\mathcal P}$ and domain adapted vocabularies are denoted as $V_{\mathcal A}$. 
WikiText~\citep{wikitext} represents the general domain that is similar to the corpus that is used for pretraining BERT, while others are chosen as the downtream domain. The red and orange bar indicate the average number of subwords tokenized with pretrain vocabulary and adapted vocabulary. We observe that AVocaDo mitigates the domain gap.

\section{Implementation Details}
\label{appendix:implementation details}

\subsection{Downstream Datasets}
\begin{table}[h]
  \begin{adjustbox}{width=0.48\textwidth}
    \begin{tabular}{l |l| l| r r r}
    \toprule
    \textbf{Domain} &\textbf{Dataset} &\textbf{Task (\# of Classes)} &\textbf{Train} & \textbf{Dev.} & \textbf{Test}
    \\ [0.1ex] 
    \midrule
    \textsc{BioMed} &\textsc{ChemProt} & relation (13) & 4169 & 2427 & 3469
    \\ [0.1ex]
    \hline  
    CS &\textsc{ACL-ARC} & citation intent (6) & 1688 &114 &139 
    \\ [0.1ex]
    \hline 
    \textsc{News} &\textsc{HyperPartisan} & partisanship (2) & 515 &65 &65 
    \\ [0.1ex]
    \hline 
    \textsc{Reviews} &\textsc{Amazon} & helpfulness (2) &115251 &5000 &25000 
    \\ [0.1ex]
    \bottomrule 
    \end{tabular}
  \end{adjustbox}
  \caption{Datasets used in experiments. Sources: \textsc{ChemProt}~\citep{10.1093/database/bav123}, ACL-ARC~\citep{jurgens-etal-2018-measuring}, \textsc{HyperPartisan}~\citep{kiesel-etal-2019-semeval},
  and \textsc{Amazon}~\citep{mcauley2015image}.}
  \label{tab:dataset}
\end{table}

We used four datasets in various domains for classification. As shown in Table~\ref{tab:dataset}, the size of the training data varies from 500 to about 110,000. The number of classes for each dataset varies from 2 to 13.

\subsection{Experimental Settings}
\label{sec:appendix:training_details}
In all experiments, we trained the networks on a single 3090 RTX GPU with 24GB of memory. We implemented all models with PyTorch using Transformers library from Huggingface. All baselines are reproduced as described in previous works~\citep{gururangan2020don,tai2020exbert,lee2020biobert,beltagy2019scibert}. 
In our experiment, the performance in \textsc{HyperPartisan} dataset tends to have high variance depending on random seeds since the size of the dataset is extremely small. 
To produce reliable results on this dataset, we discard and resample seeds.

The embeddings of newly added words in AVocaDo are initialized as a mean value of BERT embeddings of subword components. 
For instance, if the word "bluetooth" is tokenized into ["blue","\#\#tooth"] with $V_{\mathcal P}$ and "bluetooth" with $V_{\mathcal A}$, we initialize the embedding of "bluetooth" with the average value of the two subword embeddings.

\subsection{Hyperparameters}
\label{sec:appendix:implementation_details}

\begin{table}[h]
\centering
  \begin{adjustbox}{width=0.48\textwidth}
    \begin{tabular}{c |c }
    \toprule
    \textbf{Hyperparameter} &\textbf{Value} 
    \\ [0.1ex] 
    \midrule
    lower bound of fragment score $\gamma$ & 3 
    \\ [0.1ex] 
    \hline
    number of added vocabulary (initial) $\alpha$ & 500 
    \\ [0.1ex] 
    \hline
    number of added vocabulary $\beta$ & 50
    \\ [0.1ex] 
    \hline
    batch size $B$ & 16 
    \\ [0.1ex] 
    \hline
    learning rate & 1e-5, 2e-5, 5e-5
    \\ [0.1ex]
    \hline  
    number of epochs & 10
    \\ [0.1ex]
    \hline 
    temperature $\tau$ & from 1.5 to 3.5 
    \\ [0.1ex]
    \hline 
     domain vocabulary size $N_{\mathcal D}$ & 10,000 
    \\ [0.1ex]
    \bottomrule 
    \end{tabular}
  \end{adjustbox}
  \caption{Hyperparameters used in experiments. We conduct grid search for finding the best hyperparameter settings.}
  \label{tab:hyperparams}
\end{table}

\noindent As shown in Table~\ref{tab:hyperparams}, we followed the hyperparameter setting in the previous work~\citep{lee2020biobert,beltagy2019scibert,gururangan2020don}. To search the value for learning and temperature $\tau$, we use grid search.

\section{Qualitative Results}
For each downstream dataset, we randomly sampled ten words that are differently tokenized by pretrained vocabulary $V_{\mathcal P}$ and by adapted vocabulary $V_{\mathcal A}$. As shown in Table~\ref{tab:appendix:random_qualitative}, subwords tokenized by $V_{\mathcal A}$ are more informative in the target domain because they preserve the semantic of a domain-specific word.

\section{Comparison with Previous Works}

\label{appendix:our}
\begin{table}[h]
  \begin{adjustbox}{width=0.48\textwidth}
    \begin{tabular}{l | c | c} 
    \toprule
    \textbf{Model} & Adaptive Pretraining & Domain-specific Corpus
    \\ [0.4ex] 
    \midrule
    AVocaDo  & $\times$ & downstream corpus only 
    \\ [0.2ex] 
    SciBERT  &$\checkmark$	& 3.17 billion words
    \\ [0.2ex] 
    BioBERT  &$\checkmark$	& 18.0 billion words
    \\ [0.2ex] 
    exBERT  &$\checkmark$	&0.9 billion words
    \\ [0.2ex] 
    \citet{gururangan2020don}  &$\checkmark$	& 7.55 billion words
    \\ [0.2ex] 
    \toprule 
    \end{tabular}
  \end{adjustbox}
  \caption{ 
  \textbf{Comparison with previous works.}
  The adaptive pretraining phase and the size of biomedical domain corpus used for domain adaptation in previous works.
  No additional training resource is needed in AVocaDo.}
  \label{tab:no_additional_resources}
\end{table}

AVocaDo does not require additional domain-specific corpus. As shown in Table~\ref{tab:no_additional_resources}, all other baseline models require an adaptive pretraining stage before fine-tuning using domain-specific corpus. In general, the corpus used for adaptive pretraining is relatively large compared to the size of downstream dataset. Therefore, most methodologies that require adaptive pretraining require large training resources.

\section{Other Baselines}
\label{sec:appendix:other baselines}
We perform additional experiments with other baseline models. 
In this experiment, we set exBERT~\citep{tai2020exbert}, which expands the pretrained vocabulary from original
BERT$_{\text{base}}$ vocabulary, and SciBERT$_{\textsc{SCIVOCAB}}$~\citep{beltagy2019scibert}, which constructs the customized vocabulary based on science and biomedical large corpora as baselines. Table~\ref{tab:appendix_main_result} shows the overall performance on \textsc{BioMed} and \textsc{CS} domains. We outperform exBERT in \textsc{BioMed} domain. In comparison with SciBERT$_{\textsc{SCIVOCAB}}$, AVocaDo shows the competitive performance.

\section{Other Pretrained Language Models}
\label{sec:appendix:other pretrained LMs}
To demonstrate the performance of AVocaDo on the other pretrained language models, we additionally conducted experiments on RoBERTa~\citep{liu2019roberta} and ELECTRA~\citep{clark2020electra}. 
Table~\ref{tab:other_plm_results} shows the overall performance on four downstream domains. 
RoBERTa and ELECTRA with AVocaDo shows the improvements on the various domains except for \textsc{News} and \textsc{BioMed} domain respectively.

\begin{table*}[h]
\centering
\begin{adjustbox}{width=0.8\textwidth}
\begin{tabular}{l|lll}
\toprule
\textbf{Domain}               & \multicolumn{1}{c}{\textbf{Word}} & \multicolumn{1}{c}{\textbf{Pretrained Vocab $V_\mathcal{P}$}}             & \multicolumn{1}{c}{\textbf{Adapted Vocab $V_\mathcal{A}$}}    \\ \midrule
\multirow{10}{*}{\textsc{BioMed}}   & epidermal        & ep, \#\#ider, \#\#mal              & epidermal             \\
                 & cetuximab        & ce, \#\#tu, \#\#xi, \#\#ma, \#\#b        & ce, \#\#tu, \#\#xi, \#\#ma, \#\#b \\
                 & lumiracoxib       & lu, \#\#mir, \#\#aco, \#\#xi, \#\#b       & lum, \#\#irac, \#\#oxib      \\
                 & peroxidation       & per, \#\#ox, \#\#ida, \#\#tion          & perox, \#\#ida, \#\#tion     \\
                 & reductase        & red, \#\#uc, \#\#tase              & reductase             \\
                 & dihydrotestosterone   & di, \#\#hy, \#\#dro, \#\#test, \#\#ost, \#\#eron & dihydro, \#\#test, \#\#osterone  \\
                 & pparalpha        & pp, \#\#ara, \#\#pl, \#\#ha           & ppar, \#\#alpha          \\
                 & sulfhydration      & sul, \#\#f, \#\#hy, \#\#dra, \#\#tion      & sulf, \#\#hydr, \#\#ation     \\
                 & glucuronidation     & g, \#\#lu, \#\#cu, \#\#ron, \#\#ida, \#\#tion  & glucuron, \#\#ida, \#\#tion    \\
                 & proliferating      & pro, \#\#life, \#\#rating            & prolifer, \#\#ating        \\ \hline
\multirow{10}{*}{\textsc{CS}} & annotation        & ann, \#\#ota, \#\#tions             & annotation            \\
                 & unsupervised       & un, \#\#su, \#\#per, \#\#vis, \#\#ed       & unsupervised           \\
                 & entails         & en, \#\#tails                  & entail, \#\#s           \\
                 & sgd           & sg, \#\#d                    & sgd                \\
                 & parser          & par, \#\#ser                   & parser              \\
                 & nlp           & nl, \#\#p                    & nlp                \\
                 & suumarization      & sum, \#\#mar, \#\#ization            & summarization           \\
                 & syntactic        & syn, \#\#ta, \#\#ctic              & syntactic             \\
                 & coreference       & core, \#\#ference                & coreference            \\
                 & ner           & ne, \#\#r                    & ner                \\ \hline
\multirow{10}{*}{\textsc{News}}       & manafort         & mana, \#\#fort                  & manafort             \\
                 & disrespectful      & di, \#\#sr, \#\#es, \#\#pe, \#\#ct, \#\#ful   & disrespect \#\#ful        \\
                 & tweet          & t, \#\#wee, \#\#t                & tweet               \\
                 & divisive         & di, \#\#vis, \#\#ive               & div, \#\#isi, \#\#ve       \\
                 & recaptcha        & rec, \#\#ap, \#\#tch, \#\#a           & recaptcha             \\
                 & brexit          & br, \#\#ex, \#\#it                & brexit              \\
                 & irreplaceable      & ir, \#\#re, \#\#pl, \#\#ace, \#\#able      & ir, \#\#re, \#\#place, \#\#able  \\
                 & supermacists       & su, \#\#pre, \#\#mac, \#\#ists          & supermacists           \\
                 & politicize        & pol, \#\#itic, \#\#ize              & politic, \#\#ize         \\
                 & gop           & go, \#\#p                    & gop                \\ \hline
\multirow{10}{*}{\textsc{Reviews}}     & telestial        & tel, \#\#est, \#\#ial              & tele, \#\#sti, \#\#al       \\
                 & rechargeminutes     & rec, \#\#har, \#\#ge, \#\#min, \#\#ute, \#\#s  & recharge, \#\#min, \#\#utes    \\
                 & verizon         & ve, \#\#riz, \#\#on               & verizon              \\
                 & thunderbolt       & thunder, \#\#bolt                & thunderbolt            \\
                 & bluetooth        & blue, \#\#tooth                 & bluetooth             \\
                 & otterbox         & otter, \#\#box                  & otterbox             \\
                 & headset         & heads, \#\#et                  & headset              \\
                 & kickstand        & kicks, \#\#tan, \#\#d              & kickstand             \\
                 & detachable        & det, \#\#ach, \#\#able              & detach, \#\#able         \\
                 & htc           & h, \#\#tc                    & htc                \\ \bottomrule
\end{tabular}
\end{adjustbox}
\caption{\textbf{Randomly sampled words that are differently tokenized by $V_{\mathcal P}$ and $V_{\mathcal A}$.}}
\label{tab:appendix:random_qualitative}
\end{table*}

\begin{table*}[h]
\label{otherbaselines}
\centering
\adjustbox{width=1\textwidth}{

    \begin{tabular}{l| l |c c |c|c c} 
    \toprule
    \textbf{Domain} &\textbf{Dataset} & \textbf{BERT}$_{\text{base}}$
    &$\textbf{BERT}_{\text{AVocaDo}}$ & \textbf{exBERT} 
    &$\textbf{SciBERT}_{\textsc{BASEVOCAB}}$$\dagger$ 
    &$\textbf{SciBERT}_{\text{AVocaDo}}$ 
    \\ [0.1ex] 
    \midrule
    \textsc{BioMed} &\textsc{ChemProt} &79.38  &\textbf{81.07} &$74.63$
    &\textbf{83.64}  &82.71
    \\ [0.1ex]
    \hline  
    \textsc{CS} &\textsc{ACL-ARC} &56.82  &\textbf{67.28} & - &70.98 &\textbf{75.02} 
    \\ [0.1ex]
    
    \bottomrule 
    \end{tabular}
}
  \caption{\textbf{Comparisons with other baselines.} The symbol $\dagger$ indicates the performance reported by \citet{beltagy2019scibert}.} 
  \label{tab:appendix_main_result}
\end{table*}

\begin{table*}[t]
\centering 
  \begin{adjustbox}{width=0.9\textwidth}
    \begin{tabular}{l| l | c c| c c} 
    \toprule
    \textbf{Domain} &\textbf{Dataset} &\textbf{RoBERTa}$_{\text{base}}$$\dagger$ &$\textbf{RoBERTa}_{\text{AVocaDo}}$ &\textbf{ELECTRA}$_{\text{base}}$ &$\textbf{ELECTRA}_{
    \text{AVocaDo}}$ 
    \\ [0.1ex] 
    \midrule
    \textsc{BioMed} &\textsc{ChemProt} &$81.9$ &$\textbf{82.8}(+0.9)$  &$\textbf{74.3}$ &$73.4(-0.9)$    \\ [0.1ex]
    \hline  
    CS 
      &ACL-ARC &$63.0$ &$\textbf{67.3}(+4.3)$ &$57.1$ &$\textbf{59.3}(+2.2)$
    \\ [0.1ex] 
    \hline  
    \textsc{News} 
    &\textsc{HyperPartisan} &$\textbf{86.6}$ &$84.5(-2.1)$ &$70.6$ &$\textbf{77.8}(+7.2)$
    \\ [0.1ex] 
    \hline  
    \textsc{Reviews} &\textsc{Amazon} &$65.1$ &$\textbf{70.8}(+5.7)$ &$66.2$ &$\textbf{{69.9}}(+3.7)$
    \\ [0.1ex] 
    
    \bottomrule 
    \end{tabular}
  \end{adjustbox}
  \caption{\textbf{Experiments on other pretrained language models.} 
  The pretrained language models (i.e., RoBERTa$_{\text{base}}$ and ELECTRA$_{\text{base}}$) are fine-tuned with or without AVocaDo. The performance improvement is represented inside the parentheses with $+$. The symbol $\dagger$ indicates the performance reported by \citet{gururangan2020don}. }
  
\label{tab:other_plm_results}
\end{table*}

\end{document}